\documentclass[preprint,12pt]{elsarticle}

\usepackage{amssymb}
\journal{the arXiv.org}

\begin{document}

\begin{frontmatter}

\title{Question Answering  in a Natural Language Understanding System Based on Object--Oriented Semantics}

\author{Yuriy Ostapov}
\address{Institute of Cybernetics of NAS of Ukraine, 40 Acad. Glushkova  avenue,  Kiev, Ukraine.
E-mail: yugo.ost@gmail.com }

\begin{abstract}

Algorithms of question answering in a computer system oriented on input and logical processing of text information are presented.
A knowledge domain under consideration is  social behavior of a person. 
A database of the system includes an internal  representation of natural language sentences and supplemental information.
The answer {\it Yes} or {\it No} is formed for a general question.
A special question containing an interrogative word or  group of interrogative words permits  to find  a subject, object, 
place, time, cause, purpose and way of action or event.
Answer generation  is based on  identification algorithms of persons, organizations, machines, things, places, and times.  
Proposed algorithms of question answering can be realized in  information systems closely connected with text processing (criminology, operation of business, medicine, document systems).
\end{abstract}

\end{frontmatter}

\section{Introduction}

A computer understanding of natural language consists in capability of a program system to translate  sentences into an {\it  internal representation} so that this system generates adequate (i.e., valid) answers to  questions to be asked by an user.  
Adequateness is that an answer (from the viewpoint of  a researcher) solves enough correctly the problem that is contained in a question.

As the internal representation  of natural language sentence  must adequately  map  semantics of this statement,
the most natural approach is  in the simulation of facts contained in the sentence using a description of {\it real objects}  as well as actions and events connected with these objects.
This approach  was realized  in the experimental system LEIBNIZ \cite{Ostapov}.

To form an answer it is necessary, in the first place, to execute  the syntax and semantic analysis of a question,
in the second place, to find an adequate solution for the problem under review.

It should be distinguished:

\begin{itemize}

\item {\it simple questions} --- when all demandable information   is contained in a database, and only it is necessary to find an answer.
\item {\it logical questions} --- when  an inference plays a leading role for  forming answers\footnote{For example, 
it is necessary to detect on the base of indirect evidence  who is a killer. }.
                                                                                    
\end{itemize}

In this paper we are limited only simple questions. The answer is formed using the database of LEIBNIZ built from facts. 
This database consists of  predicates describing  behavior of  persons, organizations, machines, things \cite{Ostapov}.  

Analogous problems in reference to a knowledge domain under consideration (social behavior of a person) are studied in \cite{Pazzan, Schank, Smith,Thompson, Velard} .
The essential distinction of our investigation is in the use of  {\bf object--oriented semantics} to represent facts from sentences.
It should be pointed out that such approach is not attached to a specific design  tool. 
Furthermore,  a real information system can be built  on the base of this technology by means of modern design tools: {\it Microsoft Visual Studio, JBuilder, Delphi,} and others.
Information in this system will be saved and processed with the help of the living database management systems: {\it Oracle, Informix, MS SQL Server, DB2}.   

The use of  {\it Visual Prolog} \cite{Prolog} in the system LEIBNIZ is explained  by  research character of this system 
and permits with comparative ease to execute the syntax and semantic analysis as well as to organize an inference.

\section{Syntax and semantic analysis of questions}

The syntax analysis of a question is based on a description of grammatical structure (see \cite{Quirk}) using {\it Backus--Naur form}\footnote{We apply the names of grammatical constructions that were determined in \cite{Ostapov}.
The square brackets are used to point to possibility of absence for an element. }.
The question can be general or special:

$\langle$question$\rangle$ ::= $\langle$general question$\rangle$ $\mid$ $\langle$special question$\rangle$\\

The general question demands the  answer {\it Yes} or    {\it No}:\\

$\langle$general question$\rangle$ ::= $\langle$verb $\rangle$ $\langle$group of subject$\rangle$[not] $\langle$the rest of  predicate$\rangle$ $\langle$construction controlled with predicate$\rangle$\\

The verb at the beginning of the question is {\it to be, to have, to do} in the present or past indefinite, the modal verb or {\it  will (would, should)}.

The special question begins with an interrogative word or   group of interrogative words and can be addressed to an object or  adverbial modifier (or their properties):\\

$\langle$special question$\rangle$ ::= [$\langle$preposition$\rangle$] $\langle$interrogative word  or group of interrogative words$\rangle$ [$\langle$group of noun $\rangle$] $ \langle$verb$\rangle$ $\langle$group of subject$\rangle$[not] $\langle$the rest of predicate$\rangle$ $\langle$construction controlled with predicate$\rangle$\\

The interrogative word is  {\it  when, where, what, whom,} and other. The group of  interrogative words is, for example, {\it  how much, how long}. 
The verb after an interrogative word(s)  is  {\it  to be, to have, to do} in the present or past indefinite, the modal verb or  {\it will (would, should)}.\\

The special question can be addressed to a subject:\\

$\langle$special question$\rangle$ ::= $\langle${\it who} or  {\it what}$\rangle$ $\langle$predicate $\rangle$ $\langle$group of objects $\rangle$ $\langle$group of adverbial modifiers$\rangle$ \\

or to a property of  subject:\\

$\langle$special question$\rangle$ ::= $\langle${\it which, what,  whose, how many, how much}$\rangle$ \\
$\langle$group of noun$\rangle$ $\langle$predicate$\rangle$ $\langle$group of objects$\rangle$ $\langle$group of adverbial modifiers$\rangle$ \\

The special question with a compound name predicate as well can be addressed   to a subject :\\

$\langle$special question$\rangle$ ::= $\langle${\it who} or  {\it what}$\rangle$  $\langle$predicate$\rangle$ $\langle$basic noun phrase$\rangle$  $\langle$group of adverbial modifiers$\rangle$ \\

The predicate after  the interrogative word  {\it who} or {\it what }  is {\it to be}  in the indefinite tense or  the modal verb + {\it be}. \\

The semantic analysis is executed after the  parser and consists in building  predicates: {\it person, organization, machine, thing, action, event, place, time,} and others \cite{Ostapov} .

\section{Identification of objects }

Algorithms of identification are used when it is necessary to form an inquiry answer.
 Identification is the comparison  of two objects (predicates)  pointed  to the predicate of action (or  event) from a question and the similar predicate of a database  to detect their sameness\footnote{It should be pointed out that 
proposed algorithms have a certain degree of credibility, which depends on concrete conditions of application.}.

Arguments of predicates will be referred to as  {\it  fields}.
The first predicate in algorithms of identification presented below corresponds to the question, and the second predicate  to the database.

\subsection{Identification of  persons }

A person is described with the predicate {\it person}:\\[10pt]
\hspace*{20pt} 1. Code of {\it person}.\\
\hspace*{20pt} 2. Designation of person.\\
\hspace*{20pt} 3. Sex.\\
\hspace*{20pt} 4. First name.\\   
\hspace*{20pt} 5. Last name.\\  
\hspace*{20pt} 6. Additional data (other names, honorary title and degree). \\
\hspace*{20pt} 7. Place of birth (code of  {\it place}).\\ 
\hspace*{20pt} 8. Nationality. \\
\hspace*{20pt} 9. Mother tongue.\\  
\hspace*{20pt} 10. Other tongues (parallel with mother).\\
\hspace*{20pt} 11. Place of residence (code of  {\it place}).\\
\hspace*{20pt} 12. Description of face.\\ 
\hspace*{20pt} 13. Description of nose.\\ 
\hspace*{20pt} 14. Description of constitution. \\
\hspace*{20pt} 15. Description of eyes. \\
\hspace*{20pt} 16. Description of hair.\\ 
\hspace*{20pt} 17. Date of birth (code of {\it time}).\\
\hspace*{20pt} 18. Stature.\\ 
\hspace*{20pt} 19. Temperament.\\
\hspace*{20pt} 20. Psychological type.\\
\hspace*{20pt} 21. Profession.\\[10pt]

The  identification  algorithm of two predicates  {\it person}:\\[10pt]
\hspace*{20pt}1. If the first predicate contains only the field  {\it first} or  {\it last name}, then we go to step 2, or else  to step 3.\\  
\hspace*{20pt}2. If the field  {\it first} (or {\it last) name} from the  first predicate coincides  accordingly with the field {\it first} (or  {\it last) name}  from the second predicate, then the objects are identical, otherwise  the objects are not identical. The algorithm is completed.\\  
\hspace*{20pt}3.  If the first predicate contains only the fields {\it first} and  {\it last name}, then we go to step 4, or else  to step 5.\\  
\hspace*{20pt}4.  If the fields {\it first} and  {\it last name} from the first predicate  coincide accordingly with the fields  {\it first}  and  {\it last name} from the second predicate, then the objects are identical, otherwise  the objects are not identical. The algorithm is completed.\\  
\hspace*{20pt}5.  If the first predicate contains  the field  {\it first} (or {\it last) name} and a property, then we go to step 6, otherwise the algorithm is completed.\\  
\hspace*{20pt}6.  If the field {\it first} (or  {\it last) name} and this property  from the first predicate coincide accordingly with the field {\it first} (or {\it last)  name} and the same property from the second predicate, then the objects are identical, otherwise  the objects are not identical. 
The algorithm is completed.\\  

Furthermore, in specific cases one can  apply  also the comparison using only the field   {\it designation of person}. 

Analogous algorithms can be built for  the predicates {\it organization, thing,} and  {\it machine}.

\subsection{Identification of  places:}

A place of action or event is described with the predicate {\it place}.\\[10pt]
\hspace*{20pt}1. Code of  {\it place}.\\
\hspace*{20pt}2. Country.\\
\hspace*{20pt}3.  Type of region ({\it state, province}).\\   
\hspace*{20pt}4.  Geographical name of region.\\
\hspace*{20pt}5.  Territorial entity({\it town, village}).\\
\hspace*{20pt}6.  Geographical name of territorial entity. \\
\hspace*{20pt}7.  Location ({\it street, square, park, line}). \\
\hspace*{20pt}8.  Name of location.\\
\hspace*{20pt}9. Construction ({\it house, theatre, station, industrial object}).\\
\hspace*{20pt}10. Name of construction.\\
\hspace*{20pt}11. Additional information for construction ({\it stairs, roof, garret, floor}).\\ 
\hspace*{20pt}12. Designation of final location ({\it apartment, hall, office, restaurant, cafe}).\\
\hspace*{20pt}13. Designation of room ({\it bathroom, bedroom, living room, kitchen})\\

The identification algorithm of two predicates   {\it place}:\\[10pt]
\hspace*{20pt}1. If the first predicate contains only the field {\it town}, then we go to step 2, or else to step 3.\\  
\hspace*{20pt}2. If  the field {\it  town} from the first predicate coincides with the field {\it town} from the second predicate, then the objects are identical, otherwise  the objects are not identical. The algorithm is completed.\\  
\hspace*{20pt}3.  If the first predicate contains only  the fields {\it town} and   {\it street}, then we go to step 4, or else  to step 5.\\  
\hspace*{20pt}4.  If  the  fields {\it town} and  {\it street}  from the first predicate coincide  accordingly with  the  fields {\it town} and  {\it street} from the second predicate, then the objects are identical, otherwise  the objects are not identical. The algorithm is completed.\\  
\hspace*{20pt}5.  If the first predicate contains the  fields {\it  town, street,} and a number of   house (or {\it name of construction}), then we go to step 6, or else  to step 7.\\  
\hspace*{20pt}6.  If   the  fields {\it town, street,} and  the number of house (or {\it name of construction})  from the first predicate coincide accordingly with the  fields  {\it town, street,} and the number of   house (or  { \it name of construction}) from the second predicate, then the objects are identical, otherwise  the objects are not identical. The algorithm is completed.\\  
\hspace*{20pt}7.  If the first predicate contains  the  fields {\it town, street,} a number of  house, and   number of  apartment, then we go to step 8, or else  to step 9.\\  
\hspace*{20pt}8.  If  the  fields  {\it town, street,} the number of  house, and  number of apartment  from the first predicate coincide accordingly with the  fields {\it town, street,} the number of  house, and  number of apartment from the second predicate, then the objects are identical, otherwise  the objects are not identical. The algorithm is completed.\\  
\hspace*{20pt}9.   If the first predicate contains  the  fields {\it  town, street,} a  number of house,  and  location in  house ({\it stairs, roof, garret, floor}), then we go to step 10, otherwise the algorithm is completed.\\ 
\hspace*{20pt}10.  If  the  fields {\it town, street}, the number of  house, and this location in  house  from the first predicate coincide accordingly with  the  fields {\it town, street,} the number of  house, and the same location in house from the second predicate, then the objects are identical, otherwise  the objects are not identical. The algorithm is completed.\\

\subsection{Identification of  times}

A time  of action or event is described with the predicate {\it time}:\\[10pt]
\hspace*{20pt}1. Code of {\it  time}.\\
\hspace*{20pt}2. Year.\\
\hspace*{20pt}3. Season ({\it  spring, summer, autumn, winter}).\\   
\hspace*{20pt}4. Month.\\  
\hspace*{20pt}5. Day in month.\\
\hspace*{20pt}6. Day of the week. \\
\hspace*{20pt}7. Holyday ({\it  New Year, Christmas, Easter, and others}).\\
\hspace*{20pt}8. Part of  day ({\it  morning, afternoon, evening, night}).\\ 
\hspace*{20pt}9. Hours.\\

The identification algorithm of two predicates {\it time}:\\[10pt]
\hspace*{20pt}1.  If the first predicate  contains only  the field {\it year}, then we go to step 2, or else to step 3.\\  
\hspace*{20pt}2. If  the field {\it  year} from the first predicate  coincides with the field {\it year} from the second predicate, then the objects are identical, otherwise  the objects are not identical. The algorithm is completed.\\  
\hspace*{20pt}3.  If the first predicate  contains  only the fields {\it year} and   {\it season} (or  {\it month}), then we go to step 4, or else to step 5.\\  
\hspace*{20pt}4.  If  the fields {\it year} and  {\it season} (or  {\it month}) from the first predicate  coincide accordingly with the fields {\it year} and  {\it season} (or {\it month}) from the second predicate, then the objects are identical, otherwise  the objects are not identical. The algorithm is completed.\\  
\hspace*{20pt}5.  If the first predicate  contains  the fields {\it  year, month,} and  {\it day in month}, then we go to step 6, or else  to step 7.\\  
\hspace*{20pt}6.  If  the fields {\it  year, month,} and  {\it day in month}  from the first predicate  coincide accordingly with the fields  {\it year, month,} and  {\it day in month}  from the second predicate, then the objects are identical, otherwise  the objects are not identical. The algorithm is completed.\\  
\hspace*{20pt}7.   If the first predicate  contains the fields  {\it year, month, day in month,}   and   {\it hours}  (or  {\it part of day}), then we go to step 8,  otherwise the algorithm is completed.\\  
\hspace*{20pt}8.  If the fields   {\it year, month, day in month,} and  {\it hours}  (or {\it part of day}) from the first predicate  coincide accordingly with  the fields  {\it year, month, day in month,} and  {\it hours}  (or  {\it part of day})  from the second predicate, then the objects are identical, otherwise  the objects are not identical. The algorithm is completed.\\

\section{Answer generation for general questions }

Answer generation depends  essentially  on  the character of action contained in a question.
Therefore, at first  we consider in more detail  the semantic classification of verbs:\\[10pt] 
{\it  JOB} --- long purposeful occupation ({\it job, sport, studies});\\
{\it  PROPEL} --- applying a force to an object;   \\
{\it  MOVE} ---  moving a body part; \\
{\it  INGEST} ---  ingesting something inside;\\
{\it  EXPEL} ---  expelling something from a subject;\\
{\it  GRASP} ---  grasping an object;\\
{\it  GO} --- displacement of a subject;\\
{\it  TRANSFER} ---  change of general relation for a subject ({\it to buy, to sell, to come into fortune});\\
{\it  FEEL} ---  perception of a subject ({\it to see, to hear, to touch});\\
{\it  MESSAGE} ---  transmission of information between a subject and object;\\
{\it  BE} ---  identity of a subject and object, existence of a subject or connection between a subject and  certain class of objects;\\
{\it  CHANGE} --- transition of a subject to another state;\\
{\it  CREATE} ---  thinking ({\it decision-making, problem-solving, prediction});\\
{\it  DO} ---  an action.\\

This classification is founded on the fundamental investigations in \cite { Schank}
(with some additions and modifications).

The predicate {\it action} is formed for the codes PROPEL, MOVE, INGEST, EXPEL, GRASP, GO, TRANSFER, BE,  DO, the predicate  {\it job} is built for the code JOB, the predicate {\it message}  is generated for the code MESSAGE, 
the predicate  {\it intelligence} corresponds to the codes FEEL, CREATE. To form the predicate  {\it event} the code CHANGE is necessary.

Consider an algorithm of  answer generation for a general question when an action in this question  is described with the predicate  {\it action}\footnote {The structure of  {\it action} is given in \cite { Ostapov}. } :\\[10pt]
\hspace*{20pt}1. If the predicate  {\it action} from the question describes a physical effect  or change of general relation (the codes  PROPEL, MOVE, INGEST, EXPEL, GRASP, GO, TRANSFER), then we go to step 2, or else  to step 3.\\  
\hspace*{20pt}2. All the predicates {\it action} that have the same values for fields  {\it semantic type of action, negation of action, tense, type of tense } as the predicate {\it action} of the question are selected from a database. 
Each such predicate  {\it action}  is compared with  the predicate {\it action} from the question to establish identity of verbs (in terms of synonyms), subjects,  direct, indirect (or prepositional) objects,
locations,  times,  purposes, and ways of actions. If a given predicate  {\it action} from the database satisfies all these conditions, then the message  {\it Yes} is displayed.
Otherwise  an inference is executed\footnote{Algorithms of inference are not expounded in this paper as reasonably complex character of this problem demands a special consideration.} .\\
\hspace*{20pt}3. If the predicate  {\it action}  from the question has the code BE, then all the predicates {\it action} that have the code BE are selected from the database. 
Each such predicate  {\it action}  is compared with  the predicate {\it action} from the question to establish identity of subjects, 
locations, and times of actions. If a  given predicate  {\it action}  from the database  satisfies all these conditions, then the message  {\it Yes} is displayed. The algorithm is completed.
Otherwise we go to step 4.\\
\hspace*{20pt}4. All the predicates  {\it   job, message, intelligence, event}  are selected from the database. 
Each such  predicate is compared with  the predicate {\it action} from the question to establish identity of subjects, 
locations, and times of actions or events. If a given predicate from the database satisfies all the conditions, then the message  {\it Yes} is displayed. The algorithm is completed.
Otherwise  the message  {\it No} is displayed. The algorithm is completed.\\

Consider else an algorithm of  answer generation for a general question when an event in this question is described with the predicate  {\it event}\footnote {The structure of  the predicate {\it event} is given in \cite {Ostapov}. }.
 In such a situation all the predicates {\it event} that have the same  {\it  scale} as the predicate {\it event} of the question are selected from the database. 
Each such predicate {\it event} is compared with the predicate {\it event} from the question to establish  identity of verbs (in terms of synonyms), subjects, 
 locations, and times of events. If a given predicate  {\it event} from the database satisfies all these conditions, then the message  {\it Yes} is displayed.
Otherwise  the message  {\it No} is displayed.\\

 When an action in a general question is described with the predicate {\it job, message} or  {\it intelligence}, an answer for such question is formed analogously. 

As regards reliability of answers for general questions, it should be mentioned that one depends on credibility of identification algorithms.

\section{Answer generation for special questions }

 Answer generation  for special questions  is determined with two factors:

\begin{itemize}

\item   the {\it  type of predicate} for an action or event in a question;
\item  an {\it interrogative word} (or  a {\it group of interrogative words}), which  points to a solvable problem.
                                                                                    
\end{itemize}

If a problem can not be solved with a selection from a database, then an inference may help in the solution of  this task. 
Furthermore, a knowledge domain under consideration (social behavior of a person) must be  extra restricted  to provide the solution of real tasks in business, criminology, medicine, and the like.

Consider an algorithm of  answer generation for a special question when an action in this question is described with the predicate  {\it action} \footnote {Considering complexity of the algorithm for forming such answer, we  give  the reductive variant having excluded some checks and  alternatives.}:\\[10pt]
\hspace*{20pt}1. If the question is addressed to a  {\it subject} of action (the interrogative word {\it  who} or {\it  what}), and the action has the code PROPEL, MOVE,  INGEST, EXPEL, GRASP, GO, TRANSFER, then we go to step 2, or else  to step 3.\\  
\hspace*{20pt}2. All the predicates {\it action} that have the same values for fields  {\it semantic type of action, negation of action, tense, type of tense } as the predicate {\it action} of the question are selected from the database. 
Each such predicate  {\it action} is compared with the predicate {\it action} from the question to establish  identity of verbs (in terms of synonyms), direct, indirect (or prepositional) objects,
locations,  times, purposes, and ways of actions. If a given predicate  {\it action} from the database satisfies all these conditions, then  the {\it subject} from this predicate is displayed.
When all the selected predicates have been checked, and there is a true answer, then the algorithm is completed. Otherwise the inference is executed. \\
\hspace*{20pt}3. If the question is addressed to a {\it  property of the subject} \footnote {It is discovered with the parser of the question.}, and the action has the code PROPEL, MOVE,  INGEST, EXPEL, GRASP, GO, TRANSFER, then we go to step 4, or else  to step 5.\\  
\hspace*{20pt}4. All the predicates {\it action} that have the same values for fields  {\it semantic type of action, negation of action, tense, type of tense } as the predicate {\it action} of the question are selected from the database. 
Each such predicate  {\it action} is compared with the predicate {\it action} from the question to establish  identity of verbs (in terms of synonyms), subjects, direct, indirect (or prepositional)  objects,
locations, times, purposes, and ways of actions. If a given predicate  {\it action} from the database  satisfies all these conditions, then  the {\it property of the subject} from this predicate is displayed.
When all the selected predicates have been checked, and there is a true answer, then the algorithm is completed. Otherwise the answer  {\it The question can not be executed} is displayed, and algorithm is completed.\\ 
\hspace*{20pt}5. If the question is addressed to a {\it  direct object}  of action (the interrogative word {\it  what} or {\it  whom}), and the action has the code PROPEL, MOVE,  INGEST, EXPEL, GRASP, GO, TRANSFER, then we go to step 6, or else   to step 7.\\  
\hspace*{20pt}6. All the predicates {\it action} that have the same values for fields  {\it semantic type of action, negation of action, tense, type of tense } as the predicate {\it action}  of the question are selected from the database. 
Each such predicate  {\it action} is compared with the  predicate {\it action} from the question to establish  identity of verbs (in terms of synonyms), subjects, indirect (or prepositional)  objects,
locations,  times,  purposes, and ways of actions. If  a given predicate  {\it action} from the database satisfies all these conditions, then the {\it direct object} from this predicate is displayed.
When all the selected predicates have been checked, and there is a true answer, then the algorithm is completed.  Otherwise the  inference is executed.\\ 
\hspace*{20pt}7. If the question is addressed to an {\it  indirect object}  of action (the interrogative word {\it  whom}), and the action has the code PROPEL, MOVE,  INGEST, EXPEL, GRASP, GO, TRANSFER, then we go to step 8, or else  to step 9.\\  
\hspace*{20pt}8. All the predicates {\it action} that have the same values for fields  {\it semantic type of action, negation of action, tense, type of tense } as the predicate {\it action}  of the question are selected from the database. 
Each such predicate  {\it action} is compared with the predicate {\it action} from the question to establish  identity of verbs (in terms of synonyms), subjects, direct  objects,
locations, times, purposes, and ways of actions. If a given predicate  {\it action} from the database satisfies all these conditions, then  the {\it indirect object} is displayed.
When all the selected predicates have been checked, and there is a true answer, then the algorithm is completed. Otherwise the inference is executed\\ 
\hspace*{20pt}9. If the question is addressed to a {\it  time}  of action (the interrogative word {\it  when}), and the action has the code PROPEL, MOVE,  INGEST, EXPEL,  GRASP, GO,TRANSFER, then we go to step 10, or else  to step 11.\\  
\hspace*{20pt}10. All the predicates {\it action} that have the same values for fields  {\it semantic type of action, negation of action, tense, type of tense } as the predicate {\it action} of the question are selected from the database. 
Each such predicate  {\it action} is compared with the  predicate {\it action} from the question to establish  identity of verbs (in terms of synonyms), subjects,  direct, indirect (or prepositional)  objects,
locations, purposes, and ways of actions. If a given predicate  {\it action} from the database satisfies all these conditions, then the {\it time} of action from this predicate is displayed.
When all the selected predicates have been checked, and there is a true answer, then the algorithm is completed. Otherwise the inference is executed.\\ 
\hspace*{20pt}11. If the question is addressed to a {\it  place}  of action (the interrogative word {\it  where}), and the action has the code PROPEL, MOVE,  INGEST, EXPEL, GRASP, GO, TRANSFER, then we go to step 12, or else  to step 13.\\  
\hspace*{20pt}12. All the predicates {\it action} that have the same values for fields  {\it semantic type of action, negation of action, tense, type of tense } as the predicate {\it action} of the question are selected from the database. 
Each such predicate  {\it action} is compared with the predicate {\it action} from the question to establish  identity of verbs (in terms of synonyms), subjects,  direct, indirect (or prepositional)  objects,
times, purposes, and ways of actions. If a given predicate  {\it action} from the database satisfies all these conditions, then  the {\it location} of action from this predicate is displayed.
When all the selected predicates have been checked, and there is a true answer, then the algorithm is completed. Otherwise the inference is executed.\\ 
\hspace*{20pt}13. If the question is addressed to  a {\it  way}  of action (the interrogative word(s) {\it  how, in that way}), and the action has the code PROPEL, MOVE,  INGEST, EXPEL, GRASP, GO, TRANSFER, then we go to step 14, or else to step 15.\\  
\hspace*{20pt}14. All the predicates {\it action} that have the same values for fields  {\it semantic type of action, negation of action, tense, type of tense } as the predicate {\it action}  of the question are selected from the database. 
Each such predicate  {\it action}  is compared with the  predicate {\it action} from the question to establish  identity of verbs (in terms of synonyms), subjects,  direct, indirect (or prepositional) objects, locations,
times, and purposes of actions. If a given predicate  {\it action} from the database satisfies all these conditions, then the {\it way} of action from this predicate is displayed.
When all the selected predicates have been checked, and there is a true answer, then the algorithm is completed. Otherwise the inference is executed.\\ 
\hspace*{20pt}15. If the question is addressed to  a {\it  purpose}  of action (the group of interrogative words {\it  what for, for what purpose}), and the action has the code PROPEL, MOVE,  INGEST, EXPEL, GRASP, GO, TRANSFER, then we go to step 16, or else the algorithm is completed .\\  
\hspace*{20pt}16. All the predicates {\it action} that have the same values for fields  {\it semantic type of action, negation of action, tense, type of tense } as the predicate {\it action} of the question are selected from the database. 
Each such predicate {\it actions} is compared with the  predicate {\it action} from the question to establish identity of verbs (in terms of synonyms), subjects,  direct, indirect (or prepositional) objects, locations,
times, and ways of actions. If a given predicate  {\it action} from the database satisfies all these conditions, then  the {\it purpose} of action from this predicate is displayed.
When all the selected predicates have been checked, and there is a true answer, then the algorithm is completed. Otherwise  the inference is executed.\\ 

When an action or event in a special question is described with the predicate {\it job, message, intelligence} or {\it event}, an answer for such question is formed analogously. 
Similar to general questions, reliability of answers for special questions depends on credibility of identification algorithms.

\section{Example (sea story)}

Consider a story: 

 {\it Mister Brown was a mate on a ship fifteen years ago. 
The captain came up to the mate one morning. 
The captain said that he heard strange voice at night. 
This voice said in his ear to sail north-west. 
The captain told the mate to sail north-west. 
One of the men saw something black in the sea the next day. 
The captain looked through glasses. He said that there is small boat there with man. 
The captain ordered the men to save the man. Soon the men reached the small boat. 
The man was fast asleep. He went on sleeping while the men took him in  boat. 
When the man was aboard the ship, the man suddenly opened his eyes. 
The man cried out loudly where  he is. 
Captain said that ship's company saved him. 
The man asked if the captain ordered to take him from small boat. 
Captain answered that he ordered to take him. 
Then the man messaged that he executed a record voyage from New York to Liverpool on small boat}.

Questions and  answers formed by the system LEIBNIZ are given in the table:

\begin{tabbing}
AAAAAAAAAAAAAAAAAAAAAAAAAAAAAAAAA\=hspace{30ex}      \kill

Was Brown a mate?                                                        \>  Yes \\
Who was a mate?                                                            \>  Brown   \\
What did the voice say?                                                  \>  to sail north-west \\
Who ordered the men?                                                    \>  captain   \\
What did the captain order?                                            \>  to save man\\
What did the men reach?                                                 \>  boat\\
What did the captain say?                                               \>  ship's company saved man\\
Who cried out loudly?                                                       \>  the man  \\
What did the man message?                                           \>  the man did voyage\\

\end{tabbing}
 
\section{Conclusion}

As mentioned in our paper  \cite{Ostapov}, we use an ontological approach to the problem of natural language understanding. 
The ontological approach proceeds from the assumption that a natural language maps the structure of the outside world. Each sentence describes facts (or fact) 
 interpreting  actions and events for  {\it real objects}. Therefore, it is necessary  to model directly these objects, actions, and events.

As a knowledge domain, we consider social behavior of a person. These problems are studied in social psychology \cite{Mayers}.
However, when we are dealing with computer technologies,  to solve real problems the knowledge domain must be pointed in more detail.
For example, it can view  the domain of criminology (the revelation of criminal offences) \cite {Manning}  or business management  (the use of informal data about organizations, persons, goods) \cite{Chase}.

The advantage of the approach considered in our paper (as compared with living information systems in business, criminology, medicine) is that extensive text information can be used 
 {\it for logical processing}\footnote{Available text information in living systems, as a rule, can not be processed in full measure  to solve  topical problems. For example, it is difficult  to  calculate a criminal on the base of indirect evidence as only a little part of information about criminal offences
  is formalized in a database. }.

By this means the technology proposed the author discovers the perspective for successful solution of problems that so far were outside computer technologies. 
First of all this  is concern of domains of human activity where at the moment formalization of data  is connected with essential system losses (criminology, business management, medicine, document systems).


\begin{thebibliography}{1}

\bibitem{Chase}
R.B.Chase, R.F.Jacobs, W.J.Aquilano,  
\newblock Operations Management for Competitive Advantage,
\newblock McGraw-Hill, 2004. 
 

\bibitem{Quirk}
R.Quirk, S.Greenbaum, G.Leech, J.Svartvik,  
\newblock A University Grammar of English,
\newblock Longman, London, 1973. 
  
\bibitem{Manning}
P.Manning, 
\newblock The technology of Policing: Crime Mapping, Information Technology, and the Rationality of Crime Control (New Perspective in Crime, Deviance, and Law),
\newblock NYU Press, 2008. 


\bibitem{Mayers}
D.G.Myers, 
\newblock Social psychology,
\newblock McGraw-Hill, 2002. 


\bibitem{Ostapov}
Yu.Ostapov, 
\newblock Object-oriented semantics of English in natural language understanding system,
\newblock arXiv:1109.5798v1[cs.CL] 27 Sep 2011. 

\bibitem{Pazzan}
M.J.Pazzani,C.Engelman, 
\newblock Knowledge based questions answering,
\newblock The MITRE Corp., ACL Anthology.    


\bibitem{Schank}
R.C.Schank, N.M.Goldman, C.J.Rieger, C.K.Riesbeck,   
\newblock Conceptual Information Processing,  
\newblock North--Holland, Amsterdam, 1975.   

\bibitem{Smith}
R.W.Smith, D.R.Hipp, A.W.Biermann,
\newblock An Architecture for Voice Dialog System Based on Prolog-Style Theorem Proving,
\newblock Computational Linguistics, vol. 21, Number 3, 1995.   

\bibitem{Thompson}
B.H.Thompson, F.B.Thompson,   
\newblock Introducing ASK,  A Simple Knowledge System,
\newblock California Institute of Technology, ACL Anthology.   

\bibitem{Velard}
P.Velardi, M.T.Pazienza, M.Fasolo,
\newblock How to Encode Semantic Knowledge: A Method for Meaning Representation and Computer-Aided Acquisition,
\newblock Computational Linguistics, vol. 17, Number 2, 1991.   

\bibitem{Prolog}
Visual Prolog Version 5.0. Language Tutorial, 
\newblock Prolog Development Center,  
\newblock Copenhagen, 1997.  



\end{thebibliography}
\end{document}